\documentclass[letterpaper]{article} 
\usepackage{aaai25}  
\usepackage{times}  
\usepackage{helvet}  
\usepackage{courier}  
\usepackage[hyphens]{url}  
\usepackage{graphicx} 
\usepackage{siunitx}
\usepackage{booktabs}
\usepackage{multirow}

\usepackage{natbib}  
\usepackage{caption} 
\usepackage{algorithm}
\usepackage{algorithmic}
\usepackage{amsmath}
\usepackage{newfloat}
\usepackage{listings}
\DeclareCaptionStyle{ruled}{labelfont=normalfont,labelsep=colon,strut=off} 
\floatstyle{ruled}
\newfloat{listing}{tb}{lst}{}
\floatname{listing}{Listing}

\title{Inter-event Interval Microscopy for Event Cameras}
\author{
    Changqing Su\textsuperscript{\rm 1}\equalcontrib,
    Yanqin Chen\textsuperscript{\rm 2}\equalcontrib, 
    Zihan Lin\textsuperscript{\rm 3},
    Zhen Cheng\textsuperscript{\rm 4},
    You Zhou\textsuperscript{\rm 5}, 
    Bo Xiong\textsuperscript{\rm 1}\thanks{Corresponding author.}, 
    Zhaofei Yu\textsuperscript{\rm 1}, 
    Tiejun Huang\textsuperscript{\rm 1}\\
}
\affiliations{
    \textsuperscript{\rm 1}National Key Laboratory for Multimedia Information Processing, Peking University, Beijing 100871, China\\
    \textsuperscript{\rm 2}Westlake Laboratory of Life Sciences and Biomedicine, Hangzhou 310024, China\\
    \textsuperscript{\rm 3}School of Automation, Hangzhou Dianzi Uniersity, Hangzhou 310018, China\\
    \textsuperscript{\rm 4}Department of Automation, Tsinghua University, Beijing, 100084, China\\  
    \textsuperscript{\rm 5}Nanjing University Medical School, Nanjing 210023, China\\ 
    
}

\usepackage{bibentry}

\begin{document}

\maketitle

\begin{abstract}
Event cameras, an innovative bio-inspired sensor, differ from traditional cameras by sensing changes in intensity rather than directly perceiving intensity and recording these variations as a continuous stream of “events”. The intensity reconstruction from these sparse events has long been a challenging problem. Previous approaches mainly focused on transforming motion-induced events into videos or achieving intensity imaging for static scenes by integrating modulation devices at the event camera acquisition end. In this paper, for the first time, we achieve event-to-intensity conversion using a static event camera for both static and dynamic scenes in fluorescence microscopy. Unlike conventional methods that primarily rely on event integration, the proposed Inter-event Interval Microscopy (IEIM) quantifies the time interval between consecutive events at each pixel. With a fixed threshold in the event camera, the time interval can precisely represent the intensity. At the hardware level, the proposed IEIM integrates a pulse light modulation device within a microscope equipped with an event camera, termed Pulse Modulation-based Event-driven Fluorescence Microscopy. Additionally, we have collected IEIMat dataset under various scenes including high dynamic range and high-speed scenarios. Experimental results on the IEIMat dataset demonstrate that the proposed IEIM achieves superior spatial and temporal resolution, as well as a higher dynamic range, with lower bandwidth compared to other methods. The code and the IEIMat dataset will be made publicly available.
\end{abstract}

%

\section{1. Introduction}
In recent years, a novel type of neuromorphic sensor, also known as an event camera, has been developed to emulate the dynamic perception capabilities of the retinal periphery. Unlike conventional cameras that capture scene intensity directly, event cameras independently detect intensity changes at each pixel and record these changes as a stream of "event". Once the intensity change at a pixel exceeds a predefined threshold, event camera outputs an event as a four-dimensional tuple including a timestamp, pixel coordinates, and event polarity. This innovative design circumvents the exposure time limitations of traditional cameras, providing extremely high temporal resolution. Additionally, compared to traditional cameras, event cameras offer advantages such as a high dynamic range and low power consumption, demonstrating the significant potential for applications in microscopy \cite{dobbie2023event}.
\begin{figure}[t]
\centering
\includegraphics[width=1.0\columnwidth]{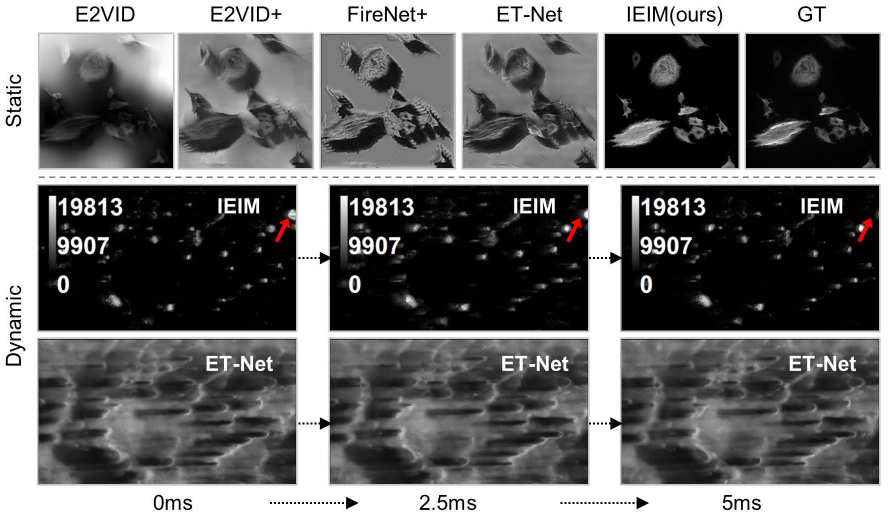} 
\caption{Event-based imaging of static and dynamic scenes. Our method achieves event-based imaging of static and dynamic scenes with quality that significantly exceeds state-of-the-art methods, offering higher temporal resolution and an extended dynamic range.}
\label{fig1}
\end{figure}
However, while the sparse event output of event cameras reduces the requirements for transmission bandwidth, it also leads to the loss of original intensity information, thus constraining their further application in certain microscopy scenarios\cite{MM2024}. To mitigate this limitation, the Dynamic and Active Pixel Vision Sensor (DAVIS) \cite{brandli2014240} was developed by incorporating an active pixel sensor into the event camera, allowing it to output both events and intensity images. By leveraging this fused data, numerous traditional computer vision tasks, such as image restoration \cite{sun2022event}, video interpolation \cite{sun2023event}, object detection and tracking \cite{zhang2021object, liu2024line,liu2024optical}, have experienced significant performance enhancements. This improvement mainly utilizes the key advantages of the event stream, such as the high temporal resolution and extensive dynamic range, to compensate for the limitations of conventional frame-based images. Nevertheless, integrating two sensors into a single chip in DAVIS leads to a concomitant reduction in sensitivity and resolution \cite{gallego2020event}, both of which are crucial for microscopic observations that typically require high sensitivity and high resolution. Some research efforts have attempted to directly convert event streams to videos. Early studies primarily relied on the intensity gradient \cite{kim2008simultaneous}, optical flow equations \cite{bardow2016simultaneous}, or on certain priors and strong assumptions \cite{munda2018real}. More recent studies \cite{rebecq2019events,cadena2021spade,stoffregen2020reducing,liang2023event} have focused on employing deep learning networks to learn the mapping from event streams to intensity images, achieving superior performance compared to earlier methods. However, these methods often suffer from detail loss, intensity distortion, and an inability to accurately reconstruct static scenes \cite{bao2024temporal}.

Therefore, the latest research \cite{bao2024temporal} has explored introducing traditional photography techniques into event cameras. By incorporating a modulation device at the acquisition end to control the continuous intensity variations on the sensor, and mapping intensity to event timestamps, researchers have achieved event-based imaging of static scenes with significant improvements over previous methods. However, this approach is inconvenient for microscopic imaging due to its severe impact on collection efficiency. The modulation at the acquisition end causes some of the fluorescence signals excited from the sample to be lost. Additionally, the modulation speed of the hardware imposes limitations on achieving high-speed dynamic imaging.

Fluorescence microscopy utilizes a light source to excite fluorescent molecules within a sample, resulting in the emission of light at specific wavelengths \cite{lichtman2005fluorescence}. This emitted fluorescence is then detected and recorded by a sensor to generate an intensity image. The differences in pixel intensity provide the structural information about the sample, which inherently arises from varying densities of fluorescent molecules in different regions of the sample. Traditional imaging methods \cite{stelzer2021light} modulate light intensity to convert densities into intensity, thereby achieving imaging of the sample. However, event stream does not contain any intensity information with the only intensity-related element being the recorded timestamp. Theoretically, the temporal resolution of an event camera is infinite, so can we directly use timestamp information to represent the densities of fluorescent molecules instead of relying on traditional intensity representation?

Based on the mentioned concept, we propose Inter-event Interval Microscopy (IEIM), a novel fluorescence microscopy technique that utilizes event cameras to capture time intervals for representing sample structure. In IEIM, the excitation light intensity of the sample is modulated in high-frequency, low-amplitude pulses rather than remaining constant. By leveraging the high-speed capability of the event camera, data is collected in the form of an event stream. The time intervals between adjacent events can reflect the structural information of the sample. We have collected the IEIMat dataset, which includes both real and simulated data from static and dynamic scenes across various samples. The results on the IEIMat dataset demonstrate that the IEIM method achieves state-of-the-art (SOTA) performance compared to existing methods, whether in static or dynamic scenes, as shown in Figure \ref{fig1}. Furthermore, compared to traditional frame-based camera methods, it offers a higher dynamic range, lower bandwidth, and higher speed. 

In summary, the main contributions of our work are three-fold:
\begin{itemize}
    \item  Based on the characteristics of fluorescence microscopy and the advantages of event cameras, we propose Inter-event Interval Microscopy (IEIM). To our knowledge, this is the first approach that directly applies event cameras to fluorescence microscopy imaging for both static and dynamic scenes.
    \item  We further collected the IEIMat dataset, which includes real and simulated data from static and dynamic scenes of various samples. We conducted a comprehensive evaluation of the proposed IEIM on the IEIMat dataset.
    \item  Results on the IEIMat dataset demonstrate the superior performance of our method. It enables fluorescence microscopy imaging of both static and dynamic scenes with higher dynamic range, lower bandwidth, and higher speed compared to traditional methods.
    
\end{itemize}

\section{2. Related Work}
\subsection{2.1 Event-based reconstruction}
Event cameras inherently lack intensity information as they only respond to intensity changes making the intensity reconstruction from event streams a widely researched problem. Numerous studies have efforts to address this problem, and their approaches can be broadly categorized into three main types: 

\textbf{(1) Traditional Methods} These methods primarily rely on gradient information provided by events, constraints from the optical flow equation, and strong assumptions or prior knowledge. For instance, in a pioneering study by Cook et al.,\cite{cook2011interacting} the brightness constancy equation constructed from events was employed for intensity reconstruction. Munda et al. \cite{munda2018real} introduced assumptions in their reconstruction approach as the uncorrelated relationship between scene structure or motion dynamics and event integration. Bardow et al. \cite{bardow2016simultaneous} proposed a reconstruction method based on variational energy minimization constraints. Additionally, some studies \cite{scheerlinck2018continuous} have also explored the direct integration of events for reconstruction, which offers good time efficiency. However, all these methods invariably suffer from severe artefacts, edge loss, and intensity distortions. 

\textbf{(2) Learning-Based Methods}. With the rapid development of deep learning, it has gradually been introduced into the field of event-based reconstruction, significantly improving performance in dynamic scenes compared to traditional methods. However, neural networks require large amounts of data for training, and obtaining such large datasets is often challenging. To address this problem, early researchers developed event simulators like ESIM \cite{rebecq2018esim} to generate events. These generated events, along with corresponding image frames, can be used to train networks. Rebecq et al. \cite{rebecq2019high} trained a convolutional neural network model, named E2VID, on synthetic data for end-to-end event-based reconstruction, greatly enhancing the quality of reconstructed videos. Cadena et al. \cite{cadena2021spade} enhanced the reconstruction details of E2VID, while Zhang et al. \cite{zhang2020learning} extended its performance in low-light scenarios. Other studies \cite{stoffregen2020reducing,weng2021event,wang2019event} have improved event-based reconstruction from different directions. Some have even achieved event-based reconstruction through self-supervised methods \cite{paredes2021back}. However, these approaches still face the reconstruction quality issues typical in traditional methods, lacking perceptual realism. Additionally, they are generally limited to events generated by motion, whether from camera or object movement. 

\textbf{(3) Photography-Based Methods}. Due to the inherent loss of intensity information in event cameras, researchers have explored to incorporate traditional photography methods during the imaging process to enhance the event camera’s imaging capabilities. He et al. \cite{he2024microsaccade} significantly improved the perceptual capability of event cameras by introducing a rotating prism at the acquisition end, which enhanced reconstruction quality to a certain extent. Bao et al. \cite{bao2024temporal} introduced a controllable aperture at the acquisition end to regulate the intensity changes perceived by the event camera, achieving high-quality imaging of static scenes with event cameras. However, these methods can adversely impact acquisition efficiency, as the modulation at the acquisition end results in the loss of some fluorescence signals excited from the sample, making them unsuitable for microscopy. Moreover, the mechanical structure of the modulation devices limits imaging speed. Therefore, developing an efficient method that enables event cameras to achieve both static and dynamic imaging in microscopy remains a challenging problem.

\subsection{2.2 High Dynamic Range Microscopy Imaging}
In fluorescence microscopy, the bit depth of the sensor determines the maximum range of detectable fluorescence signals, defining the minimum and maximum intensities that can be detected. However, the distribution and concentration of fluorescent protein expression in biological samples can vary across several orders of magnitude \cite{vinegoni2018high}. This difference can cause the actual dynamic range of the excited fluorescence to exceed the sensor's dynamic range. For instance, in neuronal imaging, the size disparity between cell bodies and neurons, as well as varying densities among cell clusters, can result in a scene with a high dynamic range \cite{vinegoni2016real}. In such scenarios, traditional imaging methods may fail, leading to indistinguishable structures in saturated areas and the loss of crucial structural information within noise. Therefore, expanding the dynamic range of imaging is crucial for fluorescence microscopy. The most intuitive method is to use a multi-exposure acquisition strategy \cite{kalantari2017deep}, where multiple images of the same scene are captured at different exposure levels and then combined using fusion algorithms to reconstruct a high dynamic range image from several low dynamic range images. This method has essentially become a standard in photography \cite{reinhard2020high}, and its basic principles have been extended to fluorescence microscopy. Vinegoni et al. \cite{vinegoni2016real} introduced a multi-exposure acquisition strategy in confocal two-photon microscopy, enabling high dynamic range imaging without additional acquisition time. However, this approach requires complex hardware modifications and faces challenges in aligning multiple low dynamic range images \cite{wang2021deep}, with the additional difficulty of achieving high-speed, high dynamic range imaging. Therefore, there is an urgent need for a method that can achieve high-speed and high dynamic range microscopy without the requirement of complex hardware modifications.

\begin{figure}[t]
\centering
\includegraphics[width=1.0\columnwidth]{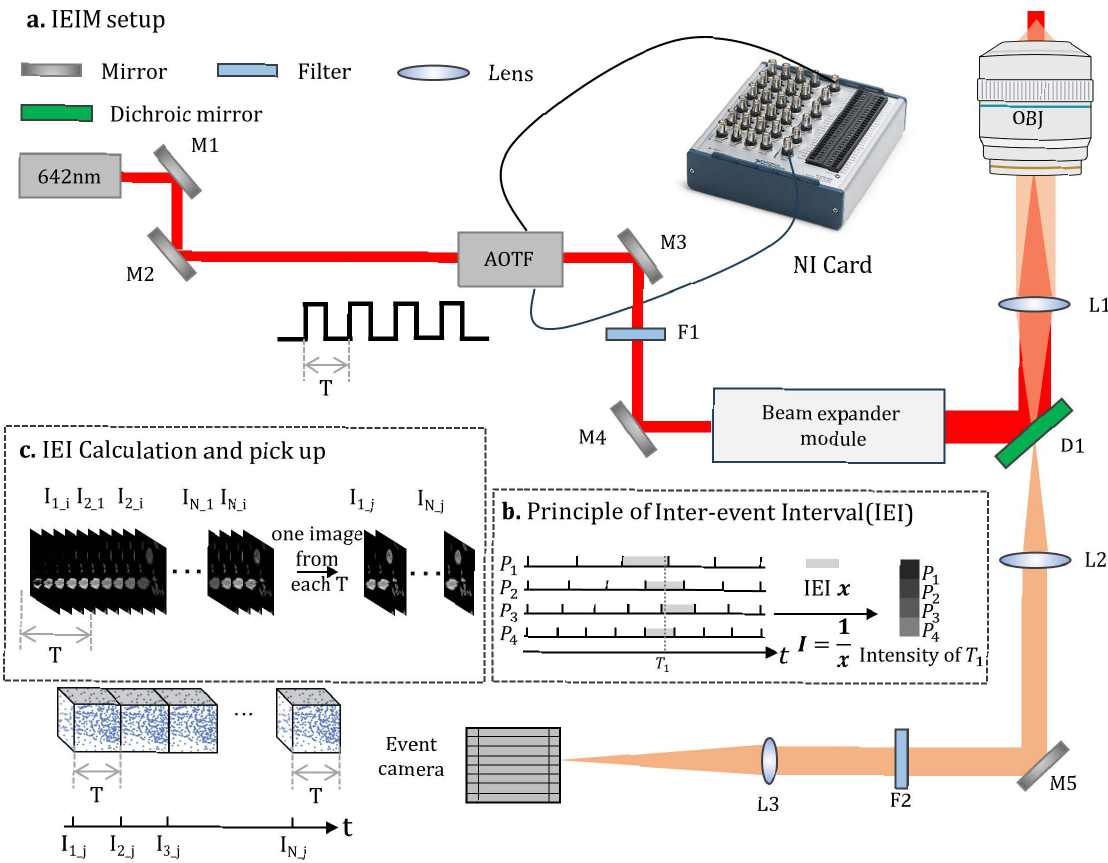} 
\caption{Pipeline of IEIM. (a) The IEIM data collection device. It employs a periodic pulsed modulation of light intensity with a period T. The collected events are processed according to the (b) Inter-Event Interval principle, where intervals between events reflect the intensity. (c) IEI calculations are performed on all event streams, and an image is selected from each cycle to represent the intensity at that specific moment.}
\label{fig2}
\end{figure}
\section{3. Inter-event Interval Microscopy}
\subsection{3.1 Events generation mechanisms}
Event cameras differ from traditional cameras in that they do not operate on a fixed frame rate. Instead, all pixels on the sensor work asynchronously, independently responding to changes in intensity on the pixel and recording these changes in the form of events. For an event $e_k$, it is represented as a four-dimensional tuple, where $e_k=(x_k, y_k, t_k, p_k)$, with $k$ indicating the $k$-th event, $(x_k, y_k)$ representing the pixel's coordinates on the sensor, and $t_k$ indicating the time the event was triggered. Once the intensity change on a pixel exceeds a predefined threshold $\theta$, an event will be triggered, which can be expressed as:
\begin{equation}
    L\left(x_k,y_k,t_k\right)-L\left(x_k,y_k,t_k-\Delta t_k\right)=\theta{p}_k
    \label{eq:1}
\end{equation}
where $L(\cdot)$ represents log-transformed pixel intensity, $\Delta t_k$ is the duration between the current event and previous event and $p_{k}\in\{+1,-1\}$ is the polarity signifying the intensity change. Here, $+1$ representing an increase in intensity and $-1$ representing a decrease. This unique design of recording and data output equips event cameras with several advantages, including high temporal resolution, low bandwidth, low latency, and a high dynamic range. These features enable high-speed recording of dynamic scene information while maintaining relatively low bandwidth.

\subsection{3.2 Principles of Inter-event Interval Microscopy}
According to the principles introduced in equation \ref{eq:1}, an event camera responds only to changes in brightness, including any factor that causes such changes. Common reconstruction algorithms, like E2VID \cite{rebecq2019high}, typically recover intensity images from events generated by motion. However, event data inherently does not record light intensity, with the only information related to intensity being the timestamp recorded in the event data. This raises the question: can we directly infer the structural information of a sample from the timestamps?  To answer this, it is essential to consider the fundamental principles of fluorescence microscopy, which relies on inconsistencies in the distribution density of fluorescent molecules that result in different intensities of fluorescence being excited by different structures within the sample. The intensity difference is merely a visualization of the densities of fluorescent molecules in the sample. Therefore, the core idea is to establish a relationship between the densities of fluorescent molecules and the event's timestamps.

Suppose that the fluorescence efficiency of a single fluorescent molecule is fixed. When the excitation light intensity suddenly changes, differences in densities of fluorescent molecules mainly affect the rate of intensity change in the sample. Event cameras, which are sensitive to these changes, can effectively capture this variation. With a fixed threshold in event camera, the rate of intensity change corresponds to the time interval between consecutive events in the event stream, or the firing frequency of the events. Consider a scenario where the initial light intensity applied to the sample is zero, meaning the sample is not yet excited. At a certain moment, a light intensity with an instantaneous increase is applied to excite the sample. Since the excitation fluorescence cannot increase instantaneously, the intensity will rise more rapidly in regions with higher densities of fluorescent molecules and more slowly in regions with lower densities. In the event records, regions where intensity changes rapidly will show smaller time intervals between consecutive events, i.e., higher firing frequencies, whereas regions with slower intensity changes will have larger time intervals, i.e., lower firing frequencies. Thus, we can directly represent the different densities of fluorescent molecules in the sample using the time intervals between consecutive events, without the need for traditional intensity reconstruction.

\subsection{3.3 Details of Implementation}
The critical component of our method is the pulse modulation of the light source, which involves varying the excitation light intensity in the form of high-frequency pulses, similar to rapidly switching the light source on and off. In our setup, an Acousto-Optic Tunable Filter (AOTF) is employed as the pulse modulation device, as illustrated in Figure \ref{fig2}. Additionally, other devices with similar functionalities, such as Electro-Optic Modulators (EOM) and Acousto-Optic Modulators (AOM), can also be employed. The AOTF functions as an electronic switch, enabling high-frequency, stable on-off operations. When integrated into the excitation light path, it modulates the excitation light to produce a pulsed variation on the sample. The temporal fluctuation in light intensity on the sample can be described by the following equation:
\begin{equation}
I(t) = \begin{cases} 
B, & \text{if } 0 \leq t \leq \frac{T}{2}, \\
0, & \text{if } \frac{T}{2} < t \leq T, 
\end{cases}
\label{eq:2}
\end{equation}
where $B$ represents the maximum excitation light intensity, and $T$ denotes the period of the pulse modulation. By adjusting the value of $T$, the acquisition speed of IEIM can be controlled, while the value of $B$ should be appropriately calibrated based on the specific sample. The intensity $B$ must exceed the light intensity threshold required to trigger an event, but it should not be excessively high. A detailed derivation will be provided below.

Suppose the fluorescence efficiency of a single fluorescent molecule is fixed at $K$, where fluorescence efficiency describes the ratio of the number of fluorescent photons emitted by the fluorophore to the number of excitation photons absorbed, the density of fluorescent molecules varies across different structural regions within a biological sample. Such variations are essential for bringing out the structural information of the sample in microscopic imaging. For the imaging sensor plane, let $D(x,y)$ represent the density of fluorescent molecules in the sample imaged at the pixel position $(x,y)$. Essentially, the aim of fluorescence imaging is to determine $D(x,y)$. Under an excitation light intensity of $B,$ the fluorescence intensity received by the imaging sensor plane in traditional imaging can be expressed as:
\begin{equation}
    m(x,y)=BKD(x,y)
\label{eq:3}
\end{equation}
where $m(x,y)$ represents the fluorescence intensity at the pixel position $(x,y)$ on the sensor plane. Since $B$ and $K$ are typically constant within a given scene, $D(x,y)$ is directly proportional to $m(x,y)$. Therefore, the intensity image recorded by the sensor effectively reflects the density of fluorescent molecules, which corresponds to the structural information of the sample. However, in event cameras, the absence of direct intensity recording makes it challenging to reconstruct high-quality structural information of the sample.

In IEIM, we introduce modulation of the excitation light intensity, so that during imaging, the light intensity no longer remains constant at $B$ but varies between $0$ and $B$ in a pulsed manner. When the light intensity abruptly shifts from $0$ to $B$, the sensor detects a continuous increase in intensity due to the bandwidth limitations of the photodiode and the fluorescence delay time, rather than an instantaneous change. Since the intensity change at the detection end is equivalent to the change at the excitation end, we can incorporate the light intensity variation into the equation, and express it as follows:
\begin{equation}
    m(x,y,t)=b(t)KD(x,y)
\label{eq:4}
\end{equation}
where $b(t)=Ht$ represents the excitation intensity that continuously increases over time $t$ with a rate $H$ and $m(x,y,t)$ denotes the fluorescence intensity at the pixel position $(x,y)$ on the sensor plane at time $t$. In event cameras, this intensity information is not recorded directly. According to the equation \ref{eq:1}, we can provide a circuit-level explanation for IEIM as follows:
\begin{equation}
    \frac{1}{C_p}\log(1+m(x,y,t))-\frac{1}{C_{p}}\log(1+m(x,y,t-1))=\theta 
\label{eq:5}
\end{equation}
where $C_p$ represents the capacitance of the photodiode, and $\frac{1}{C_p}\log(1+m(x,y,t)) = L(x,y,t)$. When the value of $m(x,y,t)$ is kept as small as possible, approaching 0, equation (5) can be simplified to:
\begin{equation}
    m(x,y,t)-m(x,y,t-1)=\theta {C_{p}}
\label{eq:6}
\end{equation}
Substituting equation \ref{eq:4} into equation \ref{eq:6} yields:
\begin{equation}
    b(t)KD(x,y)-b(t-1)KD(x,y)=\theta {C_{p}}
\label{eq:7}
\end{equation}
In event stream, $t$ and $t-1$ correspond to adjacent event timestamps $t_{(x,y,k)}$ and $t_{(x,y,(k-1))}$ at pixel position $(x,y)$, where $t_{(x,y,k)}$ denotes the timestamp of the $k$th event at pixel position $(x,y)$ and $t_{(x,y,(k-1))}$ denotes the timestamp of the $(k-1)$th event at pixel position $(x,y)$. Substituting $t_{(x,y,k)}$ and $t_{(x,y,(k-1))}$ to the $b(t)=Ht$, equation \ref{eq:7} can be further expressed as:
\begin{equation}
   Ht_{(x,y,k)}KD(x,y)- Ht_{(x,y,k-1)}KD(x,y)=\theta {C_{p}}
\label{eq:8}
\end{equation}
It can be further written as:
\begin{equation}
   D(x,y)= \frac{\theta {C_{p}}} {HK(t_{(x,y,k)}-t_{(x,y,k-1)})} 
\label{eq:9}
\end{equation}

Since $\theta$, $C_{p}$, $H$ and $K$ are constants, the density of fluorescent molecules $D(x,y)$ is inversely proportional to the time interval between adjacent events at position $(x,y)$. Thus, the time interval between adjacent events can be used to represent the structural information of the sample. 

According to the assumptions underlying equation \ref{eq:6}, IEIM requires the light intensity to fluctuate at high frequencies near 0. When the light intensity is relatively high, that is, $m(x,y,t) >> 1$, equation \ref{eq:5} is no longer approximately equal to equation \ref{eq:5} but instead approximates the following expression:
\begin{equation}
    \log(m(x,y,t))-log(m(x,y,t-1))=\theta {C_p}
\label{eq:10}
\end{equation}
Substituting equation \ref{eq:4} into equation \ref{eq:10}, it can be rewritten as:
\begin{equation}
    \log(\frac{Ht_{(x,y,k)}KD(x,y)} {Ht_{(x,y,k-1)}KD(x,y)})=\log(\frac{t_{(x,y,k)}} {t_{(x,y,k-1)}}) =\theta {C_p}
\label{eq:11}
\end{equation}
According to the equation \ref{eq:11}, the event data in this scenario no longer correlates with the density of fluorescent molecules $D(x,y)$ within the sample, but instead conforms to a fixed event generation pattern. Therefore, in fluorescence microscopy, to directly extract structural information from the event stream, it is essential to modulate the excitation light intensity with high-frequency fluctuations around 0.

\begin{figure}[b]
\centering
\includegraphics[width=1.0\columnwidth]{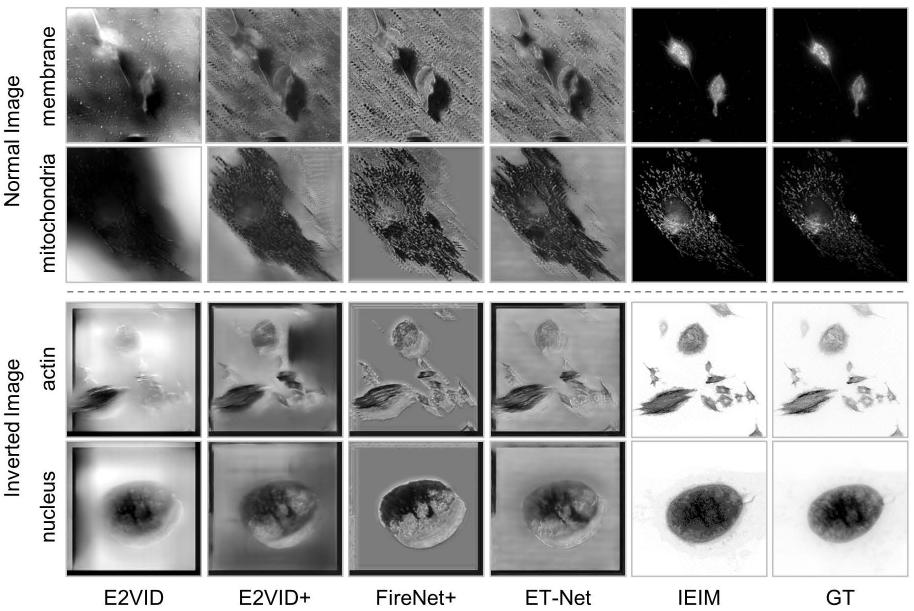} 
\caption{Comparison in synthesized static scenes. Our method significantly outperforms SOTA methods.}
\label{fig3}
\end{figure}

\begin{figure}[t]
\centering
\includegraphics[width=0.9\columnwidth]{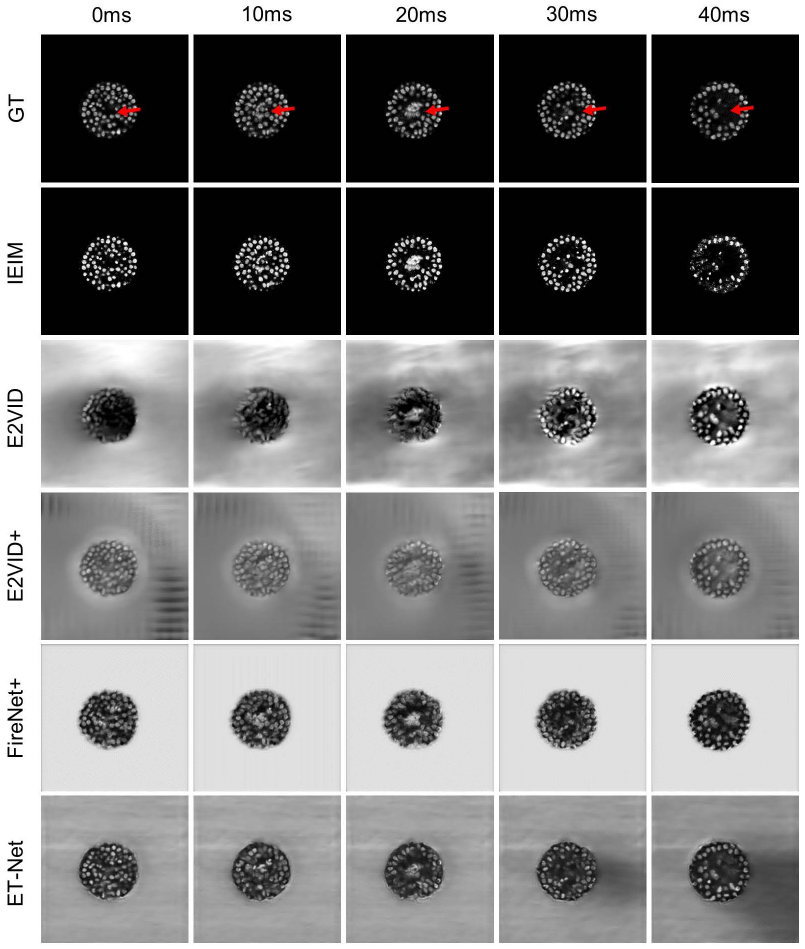} 
\caption{Comparison in synthesized motion scenes. Our method significantly outperforms SOTA methods.}
\label{fig4}
\end{figure}

\begin{table*}[ht]
\centering

\label{tab:model_comparison}
\begin{center}
\resizebox{\textwidth}{!}{ 
\begin{tabular}{l *{5}{ccc}}
\toprule
\multirow{2}{*}{Data} & \multicolumn{3}{c}{E2VID} & \multicolumn{3}{c}{E2VID+} & \multicolumn{3}{c}{ET-Net} & \multicolumn{3}{c}{FireNet+} & \multicolumn{3}{c}{IEIM (ours)} \\
\cmidrule(lr){2-4} \cmidrule(lr){5-7} \cmidrule(lr){8-10} \cmidrule(lr){11-13} \cmidrule(lr){14-16}
 & MSE$\downarrow$ & SSIM$\uparrow$ & LPIPS$\downarrow$ & MSE$\downarrow$ & SSIM$\uparrow$ & LPIPS$\downarrow$ & MSE$\downarrow$ & SSIM$\uparrow$ & LPIPS$\downarrow$ & MSE$\downarrow$ & SSIM$\uparrow$ & LPIPS$\downarrow$ & MSE$\downarrow$ & SSIM$\uparrow$ & LPIPS$\downarrow$ \\
\midrule
mem(norm\_s) & 0.224 & 0.072 & 0.607 & 0.153 & 0.059 & 0.689 & 0.236 & 0.050 & 0.748 & 0.247 & 0.053 & 0.779 & \textbf{0.001} & \textbf{0.888} & \textbf{0.152} \\
mit(norm\_s) & 0.199 & 0.182 & 0.550 & 0.191 & 0.153 & 0.630 & 0.192 & 0.160 & 0.617 & 0.218 & 0.145 & 0.603 & \textbf{0.002} & \textbf{0.838} & \textbf{0.139} \\
act(ivert\_s) & 0.153 & 0.652 & 0.475 & 0.249 & 0.583 & 0.541 & 0.160 & 0.645 & 0.473 & 0.141 & 0.593 & 0.486 & \textbf{0.003} & \textbf{0.907} & \textbf{0.162} \\
nuc(ivert\_s) & 0.177 & 0.652 & 0.450 & 0.243 & 0.604 & 0.560 & 0.141 & 0.691 & 0.496 & 0.177 & 0.642 & 0.504 & \textbf{0.003} & \textbf{0.748} & \textbf{0.343} \\
byn(norm\_d) & 0.514 & 0.037 & 0.811 & 0.245 & 0.035 & 0.653 & 0.323 & 0.048 & 0.837 & 0.654 & 0.037 & 0.665 & \textbf{0.007} & \textbf{0.915} & \textbf{0.106} \\
\bottomrule
\end{tabular}}
\end{center}
\caption{Comparison of methods on Different Metrics. The best-performing result in each group is highlighted in bold. "mem" stands for the first three letters of the sample name, "norm" indicates normal images, "ivert" refers to inverted color images, "s" denotes static scenes, and "d" signifies dynamic scenes. For example, "mem(norm\_s)" denotes a normal static image of the sample "membrane".}
\end{table*}
\section{4. Experiments}
\subsection{4.1 IEIMat Dataset}
\textbf{Synthesized data.} 
To validate the theoretical performance of the IEIM method, we collected both static and dynamic fluorescence microscopy images. The static images included actin, membrane, mitochondria, and nucleus images from \cite{hagen2021fluorescence}, while the dynamic images included Byn protein, sourced from \cite{keenan2022dynamics}. These datasets were used to simulate time series modulated by pulsed light, which were then fed into the event simulator DVS-Voltmeter \cite{lin2022dvs} to generate synthetic event stream. Since methods for network reconstruction are typically based on events generated by motion, we also generated time series by periodically shaking the images without light modulation. These sequences were similarly input into the DVS-Voltmeter to generate event stream for network reconstruction.\\
\textbf{Real-world data.} The real-world data were recorded on a custom-designed setup built around an Olympus IX-73 microscope stand and the schematic of the optical setup is presented in Figure \ref{fig2}. The excitation was performed using a continuous-wave laser at a wavelength of 642nm. The excitation will pass through an AOTF (AA Opto-Electronics, AOTFnC-400.650-TN) and an excitation filter for power control and cleaning, respectively.  The illumination modulation was realized via a data acquisition card which sends a voltage signal to the blanking channel of the AOTF to control the output power. The fluorescence collection was carried out using an Olympus 100× 1.5NA UPLAPO100XOHR oil immersion objective and imaged onto the event camera (EVK4 HD, PROPHESEE) or the referencing sCMOS camera (Hamamatsu, Fusion-BT) by relayed lens module including an emission filter (ET700/75, Chroma).

\subsection{4.2 Comparisons with State-of-the-Art Methods}
We have successfully achieved event-based fluorescence microscopy imaging for both static and dynamic scenes, a capability that currently has no equivalent in microscopy. In contrast, several macroscopic methods, such as E2VID \cite{rebecq2019high}, E2VID+ \cite{stoffregen2020reducing}, FireNet+ \cite{stoffregen2020reducing}, and ET-Net \cite{weng2021event},  can convert motion-generated events to intensity.  Therefore, we compared our method with these approaches using synthesized data. For the comparison, we utilized synthesized data without light modulation and employed pre-trained public models. The quantitative comparison results are presented in Table 1, demonstrating that our method significantly outperforms others, yielding results closest to the original images. Our method achieved an improvement of over 97\% in MSE, at least 20\% in LPIPS, and more than 14\% in SSIM. The relatively lower improvement in the SSIM metric can be attributed to the intrinsic background signal filtering characteristic of IEIM, thereby limiting the extent of enhancement. However, background signals in fluorescence microscopy are typically considered as noise, so this filtering is advantageous for fluorescence microscopy imaging.

Qualitative comparison results on synthesized data are shown in Figures \ref{fig3} and \ref{fig4}, including static and dynamic scenes, respectively. In both scenarios, our method achieved the most accurate grayscale reconstruction. Additionally, due to the differences between macroscopic and fluorescence microscopy imaging techniques, we applied a color reverse to the fluorescence images to make them more suitable for comparison with SOTA models. The results, as seen in the actin and nucleus samples in Figure \ref{fig3}, further demonstrate that our method significantly outperforms SOTA methods. Current SOTA methods exhibit intensity distortions in the results, with the most critical issue being the completely inaccurate reconstruction of background intensity. More comparison results can be find in the supplementary materials.

\subsection{4.3 IEIM in Static Imaging}
\begin{figure}[t]
\centering
\includegraphics[width=1.0\columnwidth]{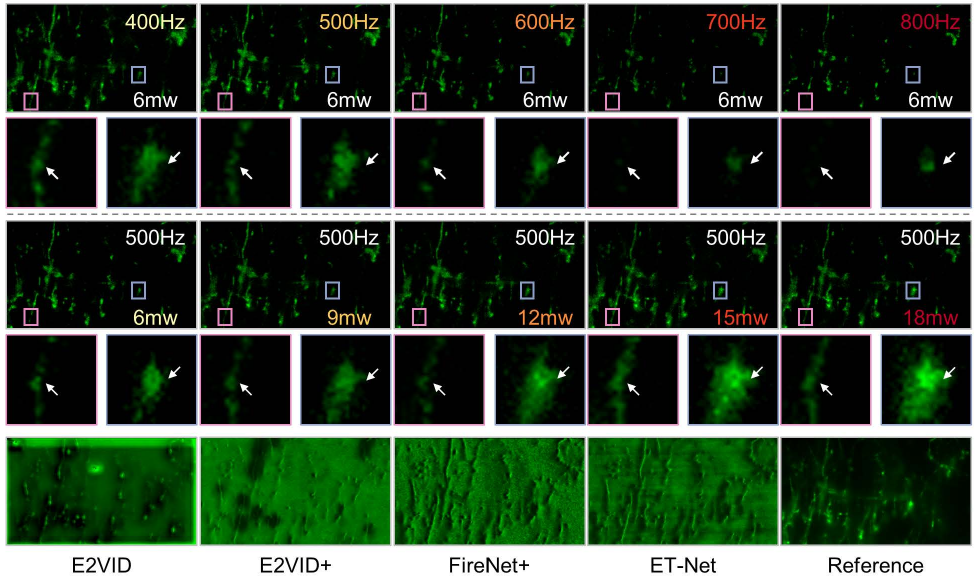} 
\caption{Comparison in real-world static scenes. Imaging with IEIM requires appropriate calibration between modulation frequency and light power and the resulting image quality far exceeds that of SOTA methods. The power in the lower right corner corresponds to the laser power at the output of the objective lens.}
\label{fig5}
\end{figure}

To evaluate the performance of the IEIM method on real-world data, we collected event streams from the sample under varying modulation frequencies and light power. To compare with SOTA methods, we also collected event streams relying solely on slow sample movement without light modulation for model reconstruction. The models used for comparison were the publicly available pre-trained models. The qualitative results are shown in Figure \ref{fig5}. As the modulation frequency increases while the light power remains constant, the imaging details of IEIM gradually decrease, as depicted in the first row of Figure \ref{fig5}, where the details in the zoomed-in images are progressively lost. This is primarily due to the event camera's sampling capability being limited by light power. when the light power is fixed, increasing the frequency shortens the exposure time, leading to insufficient data capture. This phenomenon is similar to the information loss in traditional cameras when the exposure time is too short. However, this issue can be mitigated by increasing the light power. As shown in the second row of Figure \ref{fig5}, when the modulation frequency is fixed and the light power is gradually increased, we observe that the imaging details increase and the dynamic range is also enhanced. This implies that achieving high-speed imaging requires matching the light power appropriately. Furthermore, when employing SOTA models for reconstruction, as shown in the last row of Figure \ref{fig5}, the reconstruction quality of our method significantly outperforms these methods which suffer from notable power distortion issues and completely inaccurate background reconstruction.

\subsection{4.4 IEIM in Dynamic Imaging}
\begin{figure}[t]
\centering
\includegraphics[width=1.0\columnwidth]{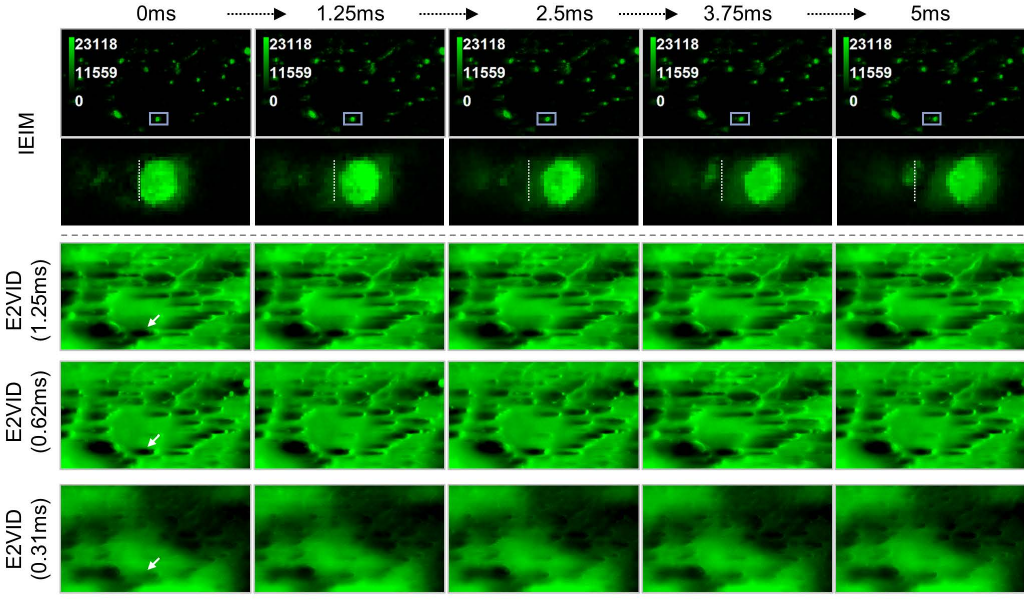} 
\caption{Comparison in real-world motion scenes. The modulation frequencies is 800Hz and light power is 9mW. Our method achieves high-quality imaging with high temporal resolution without the trade-off between temporal resolution and reconstruction detail seen in methods like E2VID. The values in parentheses indicate the temporal resolution of the reconstruction.}
\label{fig6}
\end{figure}

To evaluate the dynamic imaging capabilities of the IEIM method, we collected event streams from rapidly moving samples. We then compared the reconstruction performance of our method with SOTA methods. Due to the limitations of the light power in our acquisition system, we collected event streams from the fast-moving samples at a maximum modulation frequency of 800 Hz. Theoretically, this frequency could be further increased with enhanced light power. Since the events generated by the fast-moving samples in dynamic scenes are similar to the data used in network training, the same data was used for both network reconstruction and our method in this experiment. As shown in Figure \ref{fig6}, our method achieves a dynamic imaging speed of 800 fps in dynamic scenes while maintaining high imaging quality. Compared to the SOTA methods, our method demonstrates superior reconstruction quality. Additionally, we compared the reconstruction performance of the network methods at different temporal resolutions. The results show that in dynamic scenes, increasing the reconstruction's temporal resolution significantly reduces motion blur, but this improvement comes at the cost of losing many details. In fluorescence microscopy imaging, the observed samples typically do not exhibit uniform global motion. Typically, only a small portion of the structure is in rapid motion, while other areas remain static or move slowly. In such cases, network reconstruction methods present a trade-off between temporal resolution and reconstruction detail. In contrast, our method enables dynamic imaging with high global temporal resolution, effectively addressing this challenge.
\subsection{4.5 Ablation Study}
\begin{figure}[t]
\centering
\includegraphics[width=1.0\columnwidth]{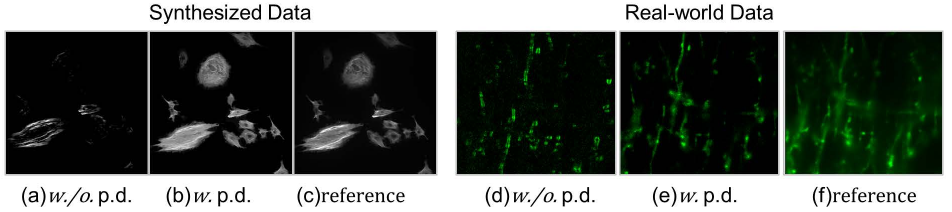} 
\caption{Visual results of ablation experiment. p.d.: pulsed modulation device. (a-c) Results in synthesized data. (d-f) Results in real-world data. }
\label{fig7}
\end{figure}

To validate the effectiveness of our modulation method, we directly applied our proposed reconstruction method to both non-modulated and modulated light intensity data, including both synthesized and real-world data. The results are shown in Figure \ref{fig7}, where it is evident that only the data with light modulation can achieve high-quality reconstruction. This is primarily because our proposed reconstruction method relies on the light intensity starting from zero and then varying in a pulsed manner. However, in the events generated by motion, the initial position already contains intensity, which does not satisfy the assumption in Equation 6 that the intensity approaches zero. As a result, the reconstruction deviates from the expected outcome.
\section{5. Conclusion}
In this paper, we have, for the first time, achieved high-quality event-based static and dynamic fluorescence microscopy imaging. We introduce the Inter-event Interval Microscopy (IEIM), which includes a uniquely designed event-based fluorescence microscopy acquisition system. This system integrates a pulsed modulation device into the excitation light path, causing the excitation light intensity on the sample to vary in a pulsed manner. Additionally, we propose a time-interval-based event reconstruction method that enables high dynamic range imaging with high temporal resolution while also achieving event-based static fluorescence microscopy imaging. To validate the performance of our method, we have collected the IEIMat dataset, which includes both synthesized and real-world data. Experiments with the IEIMat dataset demonstrate the state-of-the-art performance of our method.

\bibliography{aaai25}

\begin{thebibliography}{36}
\providecommand{\natexlab}[1]{#1}

\bibitem[{Bao et~al.(2024)Bao, Sun, Ma, and Wang}]{bao2024temporal}
Bao, Y.; Sun, L.; Ma, Y.; and Wang, K. 2024.
\newblock Temporal-Mapping Photography for Event Cameras.
\newblock \emph{arXiv preprint arXiv:2403.06443}.

\bibitem[{Bardow, Davison, and Leutenegger(2016)}]{bardow2016simultaneous}
Bardow, P.; Davison, A.~J.; and Leutenegger, S. 2016.
\newblock Simultaneous optical flow and intensity estimation from an event camera.
\newblock In \emph{Proceedings of the IEEE conference on computer vision and pattern recognition}, 884--892.

\bibitem[{Brandli et~al.(2014)Brandli, Berner, Yang, Liu, and Delbruck}]{brandli2014240}
Brandli, C.; Berner, R.; Yang, M.; Liu, S.-C.; and Delbruck, T. 2014.
\newblock A 240$\times$ 180 130 db 3 $\mu$s latency global shutter spatiotemporal vision sensor.
\newblock \emph{IEEE Journal of Solid-State Circuits}, 49(10): 2333--2341.

\bibitem[{Cadena et~al.(2021)Cadena, Qian, Wang, and Yang}]{cadena2021spade}
Cadena, P. R.~G.; Qian, Y.; Wang, C.; and Yang, M. 2021.
\newblock Spade-e2vid: Spatially-adaptive denormalization for event-based video reconstruction.
\newblock \emph{IEEE Transactions on Image Processing}, 30: 2488--2500.

\bibitem[{Cook et~al.(2011)Cook, Gugelmann, Jug, Krautz, and Steger}]{cook2011interacting}
Cook, M.; Gugelmann, L.; Jug, F.; Krautz, C.; and Steger, A. 2011.
\newblock Interacting maps for fast visual interpretation.
\newblock In \emph{The 2011 International Joint Conference on Neural Networks}, 770--776. IEEE.

\bibitem[{Dobbie(2023)}]{dobbie2023event}
Dobbie, I.~M. 2023.
\newblock Event-based super-resolution microscopy.
\newblock \emph{Nature Photonics}, 17(12): 1028--1030.

\bibitem[{Gallego et~al.(2020)Gallego, Delbr{\"u}ck, Orchard, Bartolozzi, Taba, Censi, Leutenegger, Davison, Conradt, Daniilidis et~al.}]{gallego2020event}
Gallego, G.; Delbr{\"u}ck, T.; Orchard, G.; Bartolozzi, C.; Taba, B.; Censi, A.; Leutenegger, S.; Davison, A.~J.; Conradt, J.; Daniilidis, K.; et~al. 2020.
\newblock Event-based vision: A survey.
\newblock \emph{IEEE transactions on pattern analysis and machine intelligence}, 44(1): 154--180.

\bibitem[{Hagen et~al.(2021)Hagen, Bendesky, Machado, Nguyen, Kumar, and Ventura}]{hagen2021fluorescence}
Hagen, G.~M.; Bendesky, J.; Machado, R.; Nguyen, T.-A.; Kumar, T.; and Ventura, J. 2021.
\newblock Fluorescence microscopy datasets for training deep neural networks.
\newblock \emph{GigaScience}, 10(5): giab032.

\bibitem[{He et~al.(2024)He, Wang, Zhou, Chen, Singh, Li, Gao, Shen, Wang, Cao et~al.}]{he2024microsaccade}
He, B.; Wang, Z.; Zhou, Y.; Chen, J.; Singh, C.~D.; Li, H.; Gao, Y.; Shen, S.; Wang, K.; Cao, Y.; et~al. 2024.
\newblock Microsaccade-inspired event camera for robotics.
\newblock \emph{Science Robotics}, 9(90): eadj8124.

\bibitem[{Kalantari, Ramamoorthi et~al.(2017)}]{kalantari2017deep}
Kalantari, N.~K.; Ramamoorthi, R.; et~al. 2017.
\newblock Deep high dynamic range imaging of dynamic scenes.
\newblock \emph{ACM Trans. Graph.}, 36(4): 144--1.

\bibitem[{Keenan et~al.(2022)Keenan, Avdeeva, Yang, Alber, Wieschaus, and Shvartsman}]{keenan2022dynamics}
Keenan, S.~E.; Avdeeva, M.; Yang, L.; Alber, D.~S.; Wieschaus, E.~F.; and Shvartsman, S.~Y. 2022.
\newblock Dynamics of Drosophila endoderm specification.
\newblock \emph{Proceedings of the National Academy of Sciences}, 119(15): e2112892119.

\bibitem[{Kim et~al.(2008)Kim, Handa, Benosman, Ieng, and Davison}]{kim2008simultaneous}
Kim, H.; Handa, A.; Benosman, R.; Ieng, S.-H.; and Davison, A.~J. 2008.
\newblock Simultaneous mosaicing and tracking with an event camera.
\newblock \emph{J. Solid State Circ}, 43: 566--576.

\bibitem[{Liang et~al.(2023)Liang, Zheng, Huang, Zhang, Chen, and Tian}]{liang2023event}
Liang, Q.; Zheng, X.; Huang, K.; Zhang, Y.; Chen, J.; and Tian, Y. 2023.
\newblock Event-diffusion: Event-based image reconstruction and restoration with diffusion models.
\newblock In \emph{Proceedings of the 31st ACM International Conference on Multimedia}, 3837--3846.

\bibitem[{Lichtman and Conchello(2005)}]{lichtman2005fluorescence}
Lichtman, J.~W.; and Conchello, J.-A. 2005.
\newblock Fluorescence microscopy.
\newblock \emph{Nature methods}, 2(12): 910--919.

\bibitem[{Lin et~al.(2022)Lin, Ma, Guo, and Wen}]{lin2022dvs}
Lin, S.; Ma, Y.; Guo, Z.; and Wen, B. 2022.
\newblock DVS-Voltmeter: Stochastic process-based event simulator for dynamic vision sensors.
\newblock In \emph{European Conference on Computer Vision}, 578--593. Springer.

\bibitem[{Liu et~al.(2024{\natexlab{a}})Liu, Guan, Shang, Liang, Yu, and Yu}]{liu2024optical}
Liu, Z.; Guan, B.; Shang, Y.; Liang, S.; Yu, Z.; and Yu, Q. 2024{\natexlab{a}}.
\newblock Optical Flow-Guided 6DoF Object Pose Tracking with an Event Camera.
\newblock In \emph{Proceedings of the 32nd ACM International Conference on Multimedia}, 6501--6509.

\bibitem[{Liu et~al.(2024{\natexlab{b}})Liu, Guan, Shang, Yu, and Kneip}]{liu2024line}
Liu, Z.; Guan, B.; Shang, Y.; Yu, Q.; and Kneip, L. 2024{\natexlab{b}}.
\newblock Line-based 6-DoF object pose estimation and tracking with an event camera.
\newblock \emph{IEEE Transactions on Image Processing}.

\bibitem[{Munda, Reinbacher, and Pock(2018)}]{munda2018real}
Munda, G.; Reinbacher, C.; and Pock, T. 2018.
\newblock Real-time intensity-image reconstruction for event cameras using manifold regularisation.
\newblock \emph{International Journal of Computer Vision}, 126(12): 1381--1393.

\bibitem[{Paredes-Vall{\'e}s and De~Croon(2021)}]{paredes2021back}
Paredes-Vall{\'e}s, F.; and De~Croon, G.~C. 2021.
\newblock Back to event basics: Self-supervised learning of image reconstruction for event cameras via photometric constancy.
\newblock In \emph{Proceedings of the IEEE/CVF Conference on Computer Vision and Pattern Recognition}, 3446--3455.

\bibitem[{Rebecq, Gehrig, and Scaramuzza(2018)}]{rebecq2018esim}
Rebecq, H.; Gehrig, D.; and Scaramuzza, D. 2018.
\newblock ESIM: an open event camera simulator.
\newblock In \emph{Conference on robot learning}, 969--982. PMLR.

\bibitem[{Rebecq et~al.(2019{\natexlab{a}})Rebecq, Ranftl, Koltun, and Scaramuzza}]{rebecq2019events}
Rebecq, H.; Ranftl, R.; Koltun, V.; and Scaramuzza, D. 2019{\natexlab{a}}.
\newblock Events-to-video: Bringing modern computer vision to event cameras.
\newblock In \emph{Proceedings of the IEEE/CVF Conference on Computer Vision and Pattern Recognition}, 3857--3866.

\bibitem[{Rebecq et~al.(2019{\natexlab{b}})Rebecq, Ranftl, Koltun, and Scaramuzza}]{rebecq2019high}
Rebecq, H.; Ranftl, R.; Koltun, V.; and Scaramuzza, D. 2019{\natexlab{b}}.
\newblock High speed and high dynamic range video with an event camera.
\newblock \emph{IEEE transactions on pattern analysis and machine intelligence}, 43(6): 1964--1980.

\bibitem[{Reinhard(2020)}]{reinhard2020high}
Reinhard, E. 2020.
\newblock High dynamic range imaging.
\newblock In \emph{Computer Vision: A Reference Guide}, 1--6. Springer.

\bibitem[{Scheerlinck, Barnes, and Mahony(2018)}]{scheerlinck2018continuous}
Scheerlinck, C.; Barnes, N.; and Mahony, R. 2018.
\newblock Continuous-time intensity estimation using event cameras.
\newblock In \emph{Asian Conference on Computer Vision}, 308--324. Springer.

\bibitem[{Stelzer et~al.(2021)Stelzer, Strobl, Chang, Preusser, Preibisch, McDole, and Fiolka}]{stelzer2021light}
Stelzer, E.~H.; Strobl, F.; Chang, B.-J.; Preusser, F.; Preibisch, S.; McDole, K.; and Fiolka, R. 2021.
\newblock Light sheet fluorescence microscopy.
\newblock \emph{Nature Reviews Methods Primers}, 1(1): 73.

\bibitem[{Stoffregen et~al.(2020)Stoffregen, Scheerlinck, Scaramuzza, Drummond, Barnes, Kleeman, and Mahony}]{stoffregen2020reducing}
Stoffregen, T.; Scheerlinck, C.; Scaramuzza, D.; Drummond, T.; Barnes, N.; Kleeman, L.; and Mahony, R. 2020.
\newblock Reducing the sim-to-real gap for event cameras.
\newblock In \emph{Computer Vision--ECCV 2020: 16th European Conference, Glasgow, UK, August 23--28, 2020, Proceedings, Part XXVII 16}, 534--549. Springer.

\bibitem[{Sun et~al.(2022)Sun, Sakaridis, Liang, Jiang, Yang, Sun, Ye, Wang, and Gool}]{sun2022event}
Sun, L.; Sakaridis, C.; Liang, J.; Jiang, Q.; Yang, K.; Sun, P.; Ye, Y.; Wang, K.; and Gool, L.~V. 2022.
\newblock Event-based fusion for motion deblurring with cross-modal attention.
\newblock In \emph{European conference on computer vision}, 412--428. Springer.

\bibitem[{Sun et~al.(2023)Sun, Sakaridis, Liang, Sun, Cao, Zhang, Jiang, Wang, and Van~Gool}]{sun2023event}
Sun, L.; Sakaridis, C.; Liang, J.; Sun, P.; Cao, J.; Zhang, K.; Jiang, Q.; Wang, K.; and Van~Gool, L. 2023.
\newblock Event-based frame interpolation with ad-hoc deblurring.
\newblock In \emph{Proceedings of the IEEE/CVF Conference on Computer Vision and Pattern Recognition}, 18043--18052.

\bibitem[{Vinegoni, Feruglio, and Weissleder(2018)}]{vinegoni2018high}
Vinegoni, C.; Feruglio, P.~F.; and Weissleder, R. 2018.
\newblock High dynamic range fluorescence imaging.
\newblock \emph{IEEE Journal of Selected Topics in Quantum Electronics}, 25(1): 1--7.

\bibitem[{Vinegoni et~al.(2016)Vinegoni, Leon~Swisher, Fumene~Feruglio, Giedt, Rousso, Stapleton, and Weissleder}]{vinegoni2016real}
Vinegoni, C.; Leon~Swisher, C.; Fumene~Feruglio, P.; Giedt, R.; Rousso, D.; Stapleton, S.; and Weissleder, R. 2016.
\newblock Real-time high dynamic range laser scanning microscopy.
\newblock \emph{Nature communications}, 7(1): 11077.

\bibitem[{Wang et~al.(2019)Wang, Ho, Yoon et~al.}]{wang2019event}
Wang, L.; Ho, Y.-S.; Yoon, K.-J.; et~al. 2019.
\newblock Event-based high dynamic range image and very high frame rate video generation using conditional generative adversarial networks.
\newblock In \emph{Proceedings of the IEEE/CVF Conference on Computer Vision and Pattern Recognition}, 10081--10090.

\bibitem[{Wang and Yoon(2021)}]{wang2021deep}
Wang, L.; and Yoon, K.-J. 2021.
\newblock Deep learning for hdr imaging: State-of-the-art and future trends.
\newblock \emph{IEEE transactions on pattern analysis and machine intelligence}, 44(12): 8874--8895.

\bibitem[{Weng, Zhang, and Xiong(2021)}]{weng2021event}
Weng, W.; Zhang, Y.; and Xiong, Z. 2021.
\newblock Event-based video reconstruction using transformer.
\newblock In \emph{Proceedings of the IEEE/CVF International Conference on Computer Vision}, 2563--2572.

\bibitem[{Xiong et~al.(2024)Xiong, Su, Lin, Chen, Zhou, Cheng, Yu, and Huang}]{MM2024}
Xiong, B.; Su, C.; Lin, Z.; Chen, Y.; Zhou, Y.; Cheng, Z.; Yu, Z.; and Huang, T. 2024.
\newblock Real-time Parameter Evaluation of High-speed Microfluidic Droplets using Continuous Spike Streams.
\newblock In \emph{Proceedings of the 32nd ACM International Conference on Multimedia}, MM '24, 6833–6841. New York, NY, USA: Association for Computing Machinery.
\newblock ISBN 9798400706868.

\bibitem[{Zhang et~al.(2021)Zhang, Yang, Fu, Wei, Yin, and Dong}]{zhang2021object}
Zhang, J.; Yang, X.; Fu, Y.; Wei, X.; Yin, B.; and Dong, B. 2021.
\newblock Object tracking by jointly exploiting frame and event domain.
\newblock In \emph{Proceedings of the IEEE/CVF International Conference on Computer Vision}, 13043--13052.

\bibitem[{Zhang et~al.(2020)Zhang, Zhang, Jiang, Zou, Ren, and Zhou}]{zhang2020learning}
Zhang, S.; Zhang, Y.; Jiang, Z.; Zou, D.; Ren, J.; and Zhou, B. 2020.
\newblock Learning to see in the dark with events.
\newblock In \emph{Computer Vision--ECCV 2020: 16th European Conference, Glasgow, UK, August 23--28, 2020, Proceedings, Part XVIII 16}, 666--682. Springer.

\end{thebibliography}

\end{document}